# Admissibility Alignment

## Chris Duffey

## Abstract

This paper introduces ***Admissibility Alignment***: a reframing of AI alignment as a property of admissible action and decision selection over distributions of outcomes under uncertainty, evaluated through the behavior of candidate policies. We present MAP-AI (Monte Carlo Alignment for Policy) as a canonical system architecture for operationalizing admissibility alignment, formalizing alignment as a probabilistic, decision-theoretic property rather than a static or binary condition.

MAP-AI, a new control-plane system architecture for aligned decision-making under uncertainty, in which alignment is enforced through Monte Carlo estimation of outcome distributions and admissibility-controlled Policy selection rather than static model-level constraints. The framework evaluates decision Policies across ensembles of plausible futures, explicitly modeling uncertainty, intervention effects, value ambiguity, and governance constraints. Alignment is assessed through distributional properties—expected utility, variance, tail risk, and probability of misalignment— rather than accuracy or ranking performance. This approach distinguishes probabilistic prediction from decision reasoning under uncertainty and provides an executable methodology for evaluating trust and alignment in enterprise and institutional AI systems. The result is a practical foundation for governing AI systems whose impact is determined not by individual forecasts, but by Policy behavior across distributions and tail events. Finally, we show how distributional alignment evaluation can be integrated into decision-making itself, yielding an admissibility-controlled action selection mechanism that alters Policy behavior under uncertainty without retraining or modifying underlying models.

## 1. Introduction

As artificial intelligence systems transition from passive prediction engines to systems that act, intervene, and optimize in the world, the question of alignment necessarily changes in kind. Predictive accuracy, ranking performance, and static safety checks are no longer sufficient to characterize system behavior once decisions propagate through uncertain environments, interact with human values, and generate irreversible outcomes. In such settings, alignment is not a property that can be asserted pointwise or certified once; it is a property that emerges from how a decision Policy behaves across distributions of possible futures. Alignment is not a property of internal cognition or belief correctness; it is a property of the external behavior induced by a decision Policy operating under uncertainty.

Where prior alignment work focuses on how AI systems think—through representation or reasoning—Admissibility Alignment addresses the orthogonal technical problem of how AI systems act, formalizing alignment at the level of action selection by determining which candidate actions and decisions are permitted to execute under uncertainty.

While non-agentic probabilistic monitors reduce incentives for self-preservation and deception, they do not eliminate the need to embed normative thresholds under deep uncertainty. Any guardrail that blocks actions based on predicted harm must choose how to trade false positives against catastrophic tail risks—an inherently value-laden decision that cannot be resolved ex ante. As a result, oracle-based approaches face a decision-theoretic uncertainty problem structurally similar to long-horizon reward optimization, albeit at a different layer. MAP-AI therefore treats alignment not as a fixed value embedding problem, but as a procedural decision-theoretic process over admissible actions, with explicit thresholds, auditability, and human governance.

Admissibility Alignment is operationalized through MAP-AI, a canonical decision control-plane architecture that uses Monte Carlo evaluation to govern decision selection without modifying underlying representation or reasoning models.

This paper advances a precise and operational claim: **once AI systems act under uncertainty, any alignment evaluation that does not simulate Policy behavior across outcome distributions is incomplete by definition**. The limitation is structural, not methodological. Metrics designed to evaluate predictions or scores—even when probabilistic—do not, by construction, capture how a system trades off expected performance, tail risk, value ambiguity, and governance constraints when selecting actions. As a result, alignment failures may remain undetected until they manifest operationally, often in rare but consequential regimes.

We formalize alignment as a probabilistic, decision-theoretic property of Policies rather than as a static attribute of models. The core object of evaluation is the distribution of outcomes induced by a Policy operating in an uncertain world under uncertain values and explicit constraints. Alignment, under this framing, is assessed through distributional characteristics—expected utility, variance, tail risk, and constraint violation probability—rather than through point estimates or binary compliance criteria. This reframing shifts alignment from a question of correctness to a question of risk engineering.

The alignment system architecture introduced in this paper operationalizes this definition through Monte Carlo–based alignment stress testing. By simulating Policy rollouts across ensembles of plausible environments and value specifications, the approach makes explicit the distribution of outcomes a system is likely to induce, including low-probability but high-impact events. Monte Carlo simulation is not introduced as a novel algorithmic contribution, but as a canonical instantiation of a more general requirement: alignment must be evaluated by examining Policy behavior under uncertainty, not inferred from isolated predictions or training-time objectives.

## 1.1 MAP-AI Alignment System Architecture

To operationalize alignment as a distributional property of decision-making under uncertainty, we introduce the MAP-AI Alignment System Architecture (Figure 1). The architecture decomposes alignment evaluation and control into three interacting components. **Part I** is a Monte Carlo uncertainty engine that generates distributions over possible futures by jointly sampling world realizations, Policy behavior, trajectory evolution, value uncertainty, and constraint realization. **Part II** performs cross-layer alignment stress testing by estimating distributional risk measures—such as constraint violation probability and tail risk—and projecting these risks across the model, Policy, constraint, and governance layers of the system. **Part III** integrates these risk signals into decision control through an admissibility filter and a champion–challenger decision loop, ensuring that only policies satisfying institutionally specified governance thresholds are eligible for execution. Together, these components close the alignment loop: uncertainty is transformed into decision-relevant constraints that directly shape which actions an agent is permitted to take. This architecture is agnostic to model internals and task domain, and is intended as a canonical system-level framing for alignment in agentic AI systems.

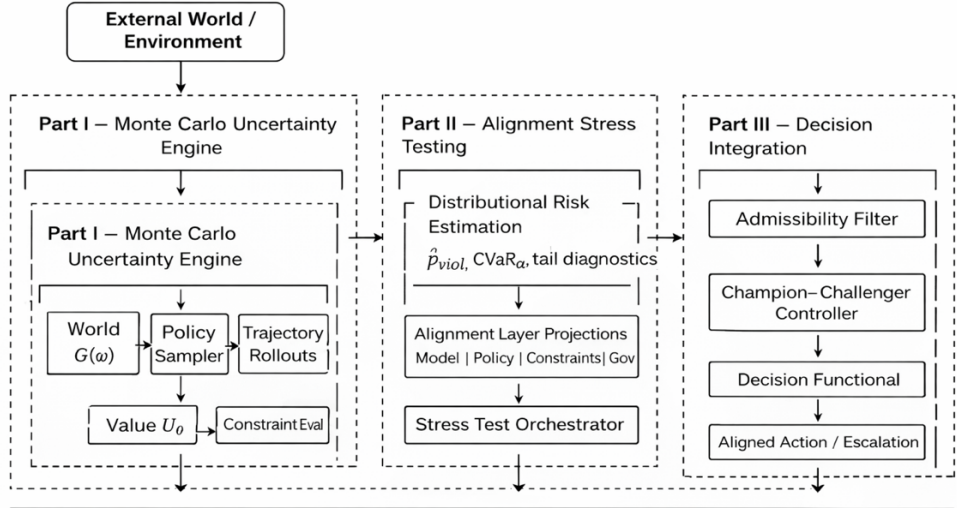

Figure 1. MAP-AI Alignment System Architecture. A canonical system-level view of MAP-AI, illustrating the separation between Monte Carlo uncertainty generation (Part I), distributional alignment stress testing (Part II), and admissibility-controlled decision integration (Part III). Alignment evaluation informs action selection without modifying model internals.

Crucially, this architecture is not limited to post hoc evaluation. Because alignment failures arise from decisions taken under uncertainty, alignment assessment must ultimately inform which actions are admissible to execute. We therefore treat alignment evaluation as an input to decision-making itself, enabling Policies to be filtered, overridden, or escalated when distributional risk exceeds institutionally specified thresholds. This distinction—between measuring alignment and enforcing it at the point of action—forms the basis of the decision integration mechanism introduced later in the paper.

**Policy (capital P)** denotes the abstract decision rule that is the object of alignment: a mapping from histories or states to actions whose induced outcome distribution is evaluated. The policy layer refers to the system-level decision interface where a Policy is instantiated and where action selection is subject to admissibility constraints. MAP-AI evaluates Policies and enforces alignment constraints at the policy layer.

**This paper makes four contributions.**

**First**, it defines alignment as a *distributional property of action selection under uncertainty*, rather than as a property of model internals, training objectives, or belief correctness.

**Second**, it introduces an **evaluation-first alignment standard** in which decision-making behavior is assessed via Monte Carlo estimation of outcome distributions, including tail risk and constraint violation probabilities, rather than optimized directly for alignment objectives.

**Third**, it formalizes **governance-admissible action selection** as a first-class construct, showing how institutional risk tolerances can be enforced through an admissibility-controlled decision interface that directly shapes which actions are executed under uncertainty.

**Fourth**, it demonstrates via stress tests that systems with equivalent predictive performance can exhibit sharply divergent alignment behavior in the tails, underscoring the insufficiency of expectation-based evaluation for alignment-critical systems.

Conceptually, this work is aligned with prior decision-theoretic treatments of AI alignment, particularly those emphasizing uncertainty over objectives and the evaluation of policies rather than predictions. Notably, Stuart Russell has argued that advanced AI systems should be understood as decision-makers operating under uncertainty about human values, and that treating objectives as fixed and known leads to systematic misalignment. This paper builds on that foundation by focusing not on how aligned policies are learned, but on how alignment can be **evaluated, compared, and governed** in deployed systems using tools already familiar to institutions that manage risk in finance, engineering, and safety-critical domains.

The contribution of this paper is therefore neither an ethical taxonomy nor a new training paradigm. It is a system-level alignment evaluation standard that treats alignment as a continuous, distributional property of decision-making systems, operationalized via Monte Carlo estimation of policy-induced outcomes. By separating alignment evaluation from capability development and optimization, the framework is intended to be complementary to existing AI systems—including forecasting and ranking platforms—while addressing a gap that becomes unavoidable as AI systems increasingly act, intervene, and optimize in the world.

## 2. Related Work

This work draws on, but is distinct from, several bodies of prior research spanning probabilistic prediction, decision-theoretic alignment, and robust evaluation under uncertainty. The common limitation across these literatures is not conceptual sophistication, but scope: most focus on *what* systems should optimize or *how* aligned behavior might be learned, rather than on *how alignment should be evaluated* once systems act in uncertain environments.

**2.1 Predictive and Ranking-Based Evaluation**

The dominant paradigm for evaluating AI systems—particularly in enterprise and institutional contexts—remains grounded in probabilistic prediction and ranking. Forecast accuracy, calibration, likelihood-based metrics, and ranking performance have proven effective for systems whose primary function is estimation rather than intervention. Even when predictions are probabilistic, evaluation typically remains pointwise, assessing whether a model assigns high probability to observed outcomes or ranks options correctly.

While these approaches are appropriate for passive inference, they do not evaluate the consequences of *acting* on predictions. Once a system selects actions that influence future states, downstream outcomes depend not only on predictive quality but on how uncertainty, tradeoffs, and constraints are resolved at decision time. As a result, alignment failures may occur despite well-calibrated predictions, particularly in rare or high-impact regimes. This limitation is structural rather than methodological: predictive metrics are not designed to characterize Policy behavior across distributions of futures.

**2.2 Decision-Theoretic Alignment and Value Uncertainty**

The closest conceptual foundation for the present work comes from decision-theoretic treatments of AI alignment that emphasize uncertainty over objectives and policies rather than fixed reward maximization. In particular, Stuart Russell has argued that advanced AI systems should be understood as decision-makers operating under uncertainty about human values, and that treating objectives as fully specified leads to systematic misalignment. Formalizations such as cooperative inverse reinforcement learning and assistance games model alignment as a cooperative decision problem in which the system must reason under incomplete information about preferences.

These frameworks establish a critical principle adopted here: alignment is fundamentally about *Policy choice under uncertainty*, not about prediction alone. However, prior work in this tradition has focused primarily on how aligned behavior might be learned or incentivized, rather than on how alignment should be evaluated in deployed systems. In particular, there is limited treatment of alignment as a continuously monitored, distributional property of outcomes under varying environmental conditions, value specifications, and governance constraints.

## 2.3 Robust and Risk-Sensitive Decision Evaluation

Adjacent to alignment research is a substantial literature on robust and risk-sensitive decision-making, including robust Markov decision processes, constrained optimization, and tail-risk–aware planning. These approaches explicitly account for uncertainty, worst-case scenarios, and risk measures such as conditional value at risk. Separately, simulation-based stress testing has been widely adopted in finance, engineering, and safety-critical systems to evaluate behavior under extreme but plausible conditions.

While these techniques provide important methodological building blocks, they are typically framed as optimization or planning tools rather than as a general evaluation standard for alignment. Moreover, they are rarely connected explicitly to questions of value uncertainty, institutional governance, or trust thresholds. As a result, their relevance to alignment is often implicit rather than formalized.

## 2.4 Summary of Gap

Taken together, prior work establishes that (i) advanced AI systems should be evaluated as decision-makers rather than predictors, and (ii) uncertainty and risk must be treated explicitly. What remains unaddressed is a unifying evaluation framework that treats alignment as a *distributional property of Policy behavior*, estimated through systematic simulation across uncertain environments, values, and constraints.

The present work fills this gap by reframing alignment evaluation as a form of decision-theoretic stress testing. Rather than proposing new objectives, learning algorithms, or ethical taxonomies, it introduces a practical methodology for comparing and governing AI decision Policies based on the distributions of outcomes they induce.

## Notation

We evaluate alignment at the level of induced **trajectory distributions** under a scenario generator, rather than model internals. The notation below defines the stochastic objects used throughout.

- **Scenario generator.** $G$ denotes a scenario generator inducing a distribution over "worlds" $\omega \sim G$. A world $\omega$ may encode environment dynamics, counterparties/adversaries, regime variables, tool availability, and other exogenous conditions relevant to deployment.
- **State, action, horizon.** A trajectory is $\tau = (s_0, a_0, s_1, a_1, \ldots, s_T)$, where $s_t \in \mathcal{S}$ is the state at time $t$, $a_t \in \mathcal{A}$ is the action at time $t$, and $T$ is the rollout horizon.
- **Policy.** A (possibly stochastic) policy $\pi$ maps decision context to actions, e.g., $a_t \sim \pi(\cdot \mid h_t)$, where $h_t$ is the information available to the Policy (state $s_t$, history, beliefs, tool outputs, and/or human inputs). MAP-AI is agnostic to how $\pi$ is implemented (model-based, model-free, hybrid, or human–AI).
- **Induced trajectory distribution.** Rolling out $\pi$ in $\omega$ (with interfaces fixed) induces a pushforward distribution over trajectories:

$$\tau \sim P(\tau \mid \pi, G)$$

This distribution is the **evaluation object**.

- **Utilities, losses, and values.** $U(\tau; \theta)$ denotes trajectory utility, with value parameters $\theta \in \Theta$ (e.g., institutional preference weights). Value uncertainty is represented by a distribution $\theta \sim P(\theta)$. When convenient, define loss $L(\tau) = -U(\tau; \theta)$ or any trajectory-level risk functional.
- **Constraints.** Let $C$ denote a set of trajectory-level constraints (e.g., prohibited actions/outcomes). Define an indicator $\mathbf{1}\{\tau \text{ violates } C\}$.
- **Monte Carlo rollouts.** MAP-AI estimates distributional quantities using i.i.d. rollouts $\tau_i \sim P(\tau \mid \pi, G)$ for $i = 1, \ldots, N$, where $N$ is the number of simulations.
- **Core estimators.**
  - Expected utility:

$$\widehat{\mathbb{E}}[U] = \frac{1}{N} \sum_{i=1}^{N} U(\tau_i)$$

- Utility variance:

$$\widehat{\text{Var}}[U] = \frac{1}{N-1} \sum_{i=1}^{N} \left(U(\tau_i) - \widehat{\mathbb{E}}[U]\right)^2$$

- Constraint violation probability:

$$\hat{p}_{\text{viol}} = \frac{1}{N} \sum_{i=1}^{N} \mathbf{1}\{\tau_i \text{ violates } C\}$$

- Tail risk ($\widehat{\text{CVaR}}_\alpha$). For $\alpha \in (0,1)$, let $\widehat{q_{1-\alpha}}$ be the empirical $1$-$\alpha$-quantile of $L(\tau)$. The empirical $\widehat{\text{CVaR}}_\alpha$ estimator is:

$$\widehat{\text{CVaR}}_\alpha = \frac{1}{|I_\alpha|} \sum_{i \in I_\alpha} L(\tau_i), I_\alpha = \{i : L(\tau_i) \geq \hat{q}_\alpha\}$$

- **Governance thresholds (admissibility).** Governance parameters $(\varepsilon, \kappa)$ define an admissible region; a Policy is admissible if, for example,

$$\hat{p}_{\text{viol}} \leq \varepsilon \text{ and } \widehat{\text{CVaR}}_\alpha \leq \kappa$$

Expected utility is compared **among admissible policies**.

- **Confidence intervals.** MAP-AI reports uncertainty using bootstrap confidence intervals (default), or asymptotic intervals where appropriate. For $\hat{p}_{\text{viol}}$ at modest $N$, Wilson or Clopper–Pearson intervals may be used.

**2.4 Relation to policy Evaluation and Risk-Sensitive Control**

MAP-AI is related to off-policy evaluation (OPE), constrained Markov decision processes (CMDPs), and risk-sensitive reinforcement learning, but differs in both objective and scope.

OPE methods typically estimate expected return or performance metrics for a fixed policy under a logged data distribution. In contrast, MAP-AI evaluates **alignment risk**, operationalized as the probability and severity of unacceptable outcomes under explicitly modeled uncertainty, including value uncertainty and governance thresholds.

Risk-sensitive and $\widehat{\text{CVaR}}_\alpha$-based RL approaches incorporate tail risk into optimization objectives. MAP-AI is intentionally **evaluation-first rather than optimization-first**: it does not prescribe how policies should be trained, nor does it assume that alignment can be achieved through objective shaping alone. Instead, it provides a common evaluation substrate for comparing policies, constraints, and governance regimes once systems are capable of acting.

Finally, MAP-AI explicitly models governance admissibility and escalation as control variables external to the policy itself, a dimension largely absent from standard CMDP and risk-sensitive control formulations.

## 3. Formal Framework

We model an AI system as selecting actions via a decision Policy in an uncertain environment with uncertain values and explicit governance constraints. Let $s_t \in \mathcal{S}$ denote the state at time $t$, $a_t \in \mathcal{A}$ the action selected, and $\omega \in \Omega$ a latent world realization capturing uncertain dynamics, regime shifts, and rare events. A scenario generator G induces a distribution $P_G(\omega)$ over such worlds.

A policy $\pi(a_t \mid s_t, h_t)$ selects actions based on state and available information $h_t$, which may include internal model state and human inputs. A trajectory $\tau = (s_0, a_0, \ldots, s_T)$ is generated by rolling out $\pi$ under world $\omega$.

Values are represented by a utility function $U(\tau; \theta)$, with uncertainty encoded via a distribution $\theta \sim P(\theta)$. Governance requirements are represented as trajectory-level constraints $g_i(\tau) \leq 0$ and an unacceptable outcome set $\mathcal{M}$.

Each Monte Carlo rollout samples jointly from world and value uncertainty, $(\omega_i, \theta_i) \sim P_G(\omega) P(\theta)$, generating an outcome trajectory $\tau_i \sim P(\tau \mid \pi, \omega_i, \theta_i)$.

The central object of alignment evaluation is the induced distribution over trajectories:

**Uncertainty decomposition.**
MAP-AI evaluates alignment by estimating the distribution of outcomes induced by a decision Policy under uncertainty. For this purpose, uncertainty is decomposed into the following components, each of which may be sampled independently or jointly in Monte Carlo evaluation:

(i) **World uncertainty**, represented by a scenario generator $G$ inducing a distribution over latent world realizations $\omega$, including environment dynamics, regime shifts, counterparties or adversaries, tool availability, and other exogenous deployment conditions;
(ii) **Policy stochasticity**, arising from randomized decision rules or stochastic inference within the policy $\pi(a_t \mid s_t, h_t)$;
(iii) **Trajectory evolution**, capturing the stochastic evolution of states and actions over a finite horizon $T$ as the policy interacts with the world;
(iv) **Value uncertainty**, represented by uncertainty over utility parameters $\theta \in \Theta$ governing institutional preferences and trade-offs; and
(v) **Constraint realization**, representing stochastic or adversarial satisfaction of trajectory-level governance constraints and unacceptable outcome sets.

This decomposition is not intended to be exhaustive of all sources of uncertainty but is sufficient to characterize the Policy-induced outcome distribution that MAP-AI treats as the alignment object.

$$P(\tau \mid \pi) = \int_\Omega \int_\Theta P(\tau \mid \pi, \omega, \theta) P_G(\omega) P(\theta) d\theta \, d\omega$$

Alignment is evaluated through distributional functionals of this distribution, including expected utility, variance, tail risk, constraint violation probability, and misalignment risk. Alignment is defined comparatively and contextually: a policy is preferred if its outcome distribution dominates ok I ho

**Table 1. Alignment Layers in MAP-AI (System-Level Framing)**

| Layer | Function | Alignment Role |
| --- | --- | --- |
| Model | Generates representations, predictions, proposals | *Pre-alignment substrate* |
| Training | Shapes tendencies, preferences, heuristics | *Weak alignment prior* |
| Policy | Selects actions under uncertainty | Alignment decision point |
| Constraints | Define admissible actions and outcomes (including evaluators and guardrails) | Alignment boundary |
| Stress Testing | Reveals distributional failures and tail behavior | Alignment visibility layer |
| Governance | Adjusts rules, thresholds, escalation, and shutdown authority | Alignment control layer |

Alignment emerges at the system level. No single layer is sufficient for alignment in isolation. Evaluator and guardrail mechanisms operate within the Constraints and Stress-Testing layers and are themselves subject to distributional evaluation.

## 4. Preemptive Caveats and Scope Conditions

This section clarifies the scope, limits, and intended interpretation of MAP-AI. The objective is to prevent common category errors—particularly the misinterpretation of MAP-AI as a training protocol, compliance checklist, or safety guarantee—by making explicit what claims the framework does and does not make. MAP-AI is not an offline or pre-deployment evaluation; this framework applies to continuous, post-deployment monitoring as policies, environments, values, and governance thresholds evolve.

**4.1 MAP-AI is an evaluation system, not a protocol**

MAP-AI is **not a protocol, checklist, or procedure whose execution guarantees alignment**. It does not prescribe how policies are constructed, trained, or optimized, nor does it specify actions that must be taken to achieve alignment. Instead, MAP-AI is a **trust and safety evaluation system** that measures, stress-tests, and governs Policy behavior under uncertainty.

The framework defines *what must be evaluated*—the distribution of outcomes induced by a Policy operating under uncertainty—and *how alignment risk should be reported*—via distributional metrics, tail risk, and governance admissibility. It does not assert that following a sequence of steps produces alignment, nor does it collapse alignment into compliance with a fixed procedure.

Alignment, under MAP-AI, is an empirical property of system behavior that must be observed, compared, and governed—not a box that can be checked.

**4.2 Conditionality of all alignment claims**

All MAP-AI results are **conditional**. Reported alignment metrics are conditional on:

- the declared scenario generator and its abstractions,
- the modeled interfaces (including human involvement and tooling),
- the specified value parameter distributions,
- and the governance thresholds used to define admissibility.

MAP-AI makes no unconditional claims about real-world frequencies or universal safety. When scenario generators or evaluators are mis-specified, results should be interpreted as *bounds under stated assumptions*, not as guarantees. This conditionality is a feature rather than a limitation: it makes assumptions explicit and auditable, rather than implicit and unverifiable.

**4.3 Evaluator and measurement model risk**

MAP-AI treats **all evaluators as fallible system components**. This includes automated constraint classifiers, harm estimators, red-teaming models, and guardrail mechanisms, whether human-operated or automated.

Evaluator-first safety architectures—such as probabilistic harm estimators or automated guardrail systems—can be incorporated within MAP-AI as implementations of constraint evaluation or gating logic **applied to policy outcomes rather than internal beliefs**. However, their outputs are treated as noisy measurements, not as ground truth. Evaluator error, calibration drift, blind spots, and normative ambiguity are therefore sources of system risk that propagate into misalignment estimates.

MAP-AI does not assume evaluator trustworthiness by design. Instead, it requires that the effect of evaluators and guardrails be assessed empirically through their impact on the induced trajectory distribution.

**4.4 Discovery versus estimation**

MAP-AI explicitly separates **failure discovery** from **risk estimation**.

Procedures that over-sample hazardous regions—such as adversarial scenario search, rare-event amplification, or evaluator-guided stress generation—are permitted and encouraged for discovering alignment-relevant failure modes. However, reported alignment metrics must always be computed under a **declared evaluation distribution**, whether the base scenario generator or a specified stress distribution.

This separation is required to avoid conflating "failures were found" with "failures are likely." MAP-AI therefore treats adversarial or evaluator-guided sampling as a discovery tool, not as a substitute for distributional estimation.

### 4.5 Thresholds, admissibility, and governance are not moral facts

Governance thresholds defining admissibility regions are **institutional decisions**, not objective truths. A policy deemed inadmissible under one set of risk tolerances may be admissible under another. MAP-AI makes these thresholds explicit and evaluates their consequences, rather than embedding them implicitly in optimization objectives or safety rules.

Thresholding introduces discontinuities: small changes in tolerances can induce large changes in the admissible policy set. MAP-AI treats this sensitivity as an object of evaluation, not as a defect. Governance is therefore modeled as an **active control layer**, not as a static compliance filter.

### 4.6 Alignment is not localized to a single layer

MAP-AI does not locate alignment in any single component of an AI system. Models generate representations and proposals; training shapes tendencies; policies select actions; constraints define admissible behavior; stress testing reveals distributional failures; governance adjusts thresholds and escalation rules.

Alignment emerges from the interaction of these layers once a system has the ability to act. Monte Carlo simulation is not itself the alignment layer; it is the **evaluation substrate** that makes the interaction between layers observable under uncertainty.

### 4.7 Human oversight is not assumed to guarantee alignment

Human involvement—approval gates, escalation rules, overrides—is treated as a **policy component**, not as a guarantee. MAP-AI evaluates whether oversight mechanisms change outcome distributions, tail risk, and admissibility in practice. It does not assume that the presence of a human in the loop ensures alignment by design.

This mirrors institutional safety practice in other domains: oversight mechanisms are evaluated by their realized effect on outcomes, not by their intent.

### 4.8 No claim of completeness or finality

MAP-AI does not claim to solve alignment, eliminate emergent risk, or replace interpretability, training-time alignment, or governance institutions. Its claim is narrower and more defensible: **once AI systems act under uncertainty, alignment cannot be meaningfully assessed without evaluating policy behavior across outcome distributions**.

Absent such evaluation, alignment remains an assumption rather than an operational property.

## 5. Monte Carlo Alignment Stress-Testing Methodology

Alignment metrics in MAP-AI are estimated via repeated simulation of Policy behavior under a declared scenario generator. This section specifies the stochastic evaluation object, the estimator family used to quantify alignment risk, the treatment of tail events and rare failures, and the calibration and reporting requirements needed to make results comparable across policies and reproducible across evaluation contexts.

### 5.1 Rollouts and the stochastic evaluation object

Let $G$ denote a scenario generator that induces a distribution over possible worlds $\omega$. For a candidate policy $\pi$, MAP-AI evaluates the induced distribution over trajectories

$$\tau = (s_0, a_0, s_1, a_1, \ldots, s_T)$$

produced by rolling out $\pi$ in $\omega$, with any required human-in-the-loop mechanisms and tool interfaces held fixed.

The evaluation object is therefore the **pushforward distribution**

$$P(\tau \mid \pi, G)$$

not the internal representations, objectives, or training dynamics of the underlying model.

MAP-AI separates sources of uncertainty by construction:

1. **Exogenous uncertainty** in $\omega$ (environmental dynamics, counterparties, market regimes),
2. **Model uncertainty** (epistemic uncertainty in beliefs or forecasts used by $\pi$),
3. **Value uncertainty** (uncertainty over utility parameters and institutional preferences).

These sources may be sampled independently or jointly. All reported results must explicitly state which components are randomized and which are conditioned.

### 5.2 Alignment metrics and estimators

Let $\{\tau_i\}_{i=1}^N \sim P(\tau \mid \pi, G)$ denote $N$ independent rollouts of Policy $\pi$. For any trajectory-level functional $f(\tau)$ (e.g., utility, loss, or constraint indicator), MAP-AI estimates distributional quantities using the following canonical estimators. Throughout this section, $\widehat{\mathrm{CVaR}}_\alpha$ is computed over loss, where the loss is defined as $L(\tau) := -U(\tau)$.

**Expected utility**

$$\widehat{\mathbb{E}}[U] = \frac{1}{N} \sum_{i=1}^{N} U(\tau_i)$$

**Dispersion (sample variance)**

$$\widehat{\mathrm{Var}}[U] = \frac{1}{N-1} \sum_{i=1}^{N} \left( U(\tau_i) - \widehat{\mathbb{E}}[U] \right)^2$$

**Constraint violation probability**

For trajectory-level constraints $C$,

$$\hat{p}_{\mathrm{viol}} = \frac{1}{N} \sum_{i=1}^{N} \mathbf{1}\{\tau_i \text{ violates } C\}$$

**Tail risk (Conditional Value-at-Risk)**

Let L(τ) denote loss, and let q_hat_(1−α) be the empirical (1−α)-quantile of L(τ). Define the index

$$I_\alpha = \{ i : L(\tau_i) \geq \hat{q}_{(1-\alpha)} \}.$$

The empirical $\widehat{\text{CVaR}}_\alpha$ estimator is

$$\widehat{\text{CVaR}}_\alpha = \frac{1}{|I_\alpha|} \sum_{i \in I_\alpha} L(\tau_i)$$

**Governance admissibility**

MAP-AI treats governance thresholds as first-class. A policy $\pi$ is **admissible** only if

$$\hat{p}_{\text{viol}} \leq \varepsilon \text{ and } \widehat{\text{CVaR}}_\alpha \leq \kappa$$

for institutionally specified tolerances $(\varepsilon, \kappa)$.

Expected utility is **not sufficient** for admissibility; it is evaluated only among admissible policies.

**Estimator uncertainty**

Uncertainty in all reported estimates must be quantified. MAP-AI supports:

- nonparametric bootstrap confidence intervals for $\hat{\mathbb{E}}[U]$, $\hat{p}_{\text{viol}}$, and $\widehat{\text{CVaR}}_\alpha$;
- asymptotic intervals when regularity conditions are satisfied;
- Wilson or Clopper–Pearson intervals for violation probabilities when $N$ is modest.

Unless otherwise stated, results are reported with 95% confidence intervals.

### 5.3 Tail discovery and rare-event amplification

Naïve Monte Carlo sampling under-represents rare but high-impact failures. MAP-AI therefore supports variance-reduction and rare-event amplification techniques as part of the evaluation protocol.

Two families are recommended:

1. **Stratified or conditional sampling**, in which rollouts are allocated across scenario strata (e.g., regimes defined by $G$) to ensure coverage of relevant operational states and enable regime-conditional risk reporting.
2. **Importance sampling**, using a proposal distribution $Q(\omega)$ that over-weights known hazard regions. When likelihood ratios are available, unbiased estimators may be constructed; otherwise, resulting estimates are treated as conservative upper bounds.

$$w(\omega) = \frac{P_G(\omega)}{Q(\omega)}$$

Adversarial scenario search can be interpreted as constructing $Q$ to target failure discovery. In MAP-AI, such procedures are permitted for **discovery**, but any reported alignment metric must be computed under a declared evaluation distribution (either the base generator $G$ or an explicitly specified stress distribution) to avoid conflating discovery with estimation.

### 5.4 Simulator calibration and model risk

All Monte Carlo results are conditional on the scenario generator $G$ and any simulator used to produce rollouts. MAP-AI treats this dependence as **model risk** and requires explicit reporting of:

1. the assumed dynamics, interfaces, and abstractions represented in $G$;
2. calibration targets and procedures, analogous to backtesting in finance;
3. sensitivity analyses over key scenario parameters.

Calibration is assessed by whether the simulator reproduces known regimes and failure modes relevant to the deployment context. When calibration is weak, MAP-AI results are interpreted as bounds under stated assumptions rather than claims about real-world frequencies.

**Scenario generator instantiations.**
In practice, the scenario generator $G$ may take several forms, including:
(i) physics- or rules-based simulators calibrated to known regimes;
(ii) learned world models trained on historical interaction data;
(iii) hybrid generators combining empirical logs with synthetic stress scenarios; and
(iv) curated adversarial or red-team scenario suites.

MAP-AI is agnostic to the specific instantiation of $G$, but requires that its abstractions, calibration targets, and known failure modes be explicitly reported. Differences in $G$ are treated as model risk rather than hidden assumptions.

### 5.5 Practical implementation notes

To support comparability across candidate policies, MAP-AI fixes the evaluation horizon $T$, applies random-seed control for paired evaluations, and enforces identical interface constraints (e.g., human oversight rules, tool access, rate limits). Where possible, policies are evaluated using common random numbers to reduce estimator variance in pairwise comparisons.

Human involvement enters only through its operationalized effect on trajectories—approval gates, escalation rules, and constraint overrides. MAP-AI evaluates these mechanisms empirically as part of the policy, rather than assuming they guarantee alignment by design.

## 6. Stress Tests

This section instantiates Part II of the MAP-AI Alignment System Architecture (alignment stress testing) in its simplest non-trivial form. To keep the diagnostic example maximally interpretable, the stress tests in this section vary a **single uncertainty dimension**—**world uncertainty via the scenario generator** $G$—while holding Policy stochasticity, value parameters, and constraint realization fixed. This isolates the effect of environmental uncertainty on alignment-relevant behavior without confounding effects from learning dynamics or preference drift.

### 6.1 Predictive parity with divergent tail behavior

Accordingly, the worked example in Section 6.1 varies only world uncertainty via the scenario generator, holding Policy stochasticity, value parameters, and constraint realization fixed. We illustrate MAP-AI using a minimal diagnostic stress test sufficient to induce distributional divergence under identical predictive beliefs.

We model uncertainty using a **regime-switching environment**, where *regime* is used in its standard technical sense to denote latent operating modes of a stochastic system, as in regime-switching Markov models, econometric bull/bear regimes, control-theoretic operating modes, and reliability models distinguishing normal and fault conditions.

### Scenario generator

The scenario generator $G$ induces world uncertainty through a latent operational regime variable with two modes: **Normal** and **Adverse**. The system begins in the Normal regime and transitions to the Adverse regime at each timestep with probability $p = 0.02$. Once entered, the Adverse regime is absorbing. The rollout horizon is fixed at $T = 20$.

This structure captures rare but persistent adverse operating conditions—such as distribution shift, cascading failures, or adversarial onset—without assuming any particular task domain, model architecture, or internal reasoning process.

### Policies

We evaluate two deterministic policies, $\pi_A$ and $\pi_B$, that share an identical predictive belief state and identical observations. Both policies assign the same probability to regime transitions and differ only in their action mapping under uncertainty.

- **Policy** $\pi_A$ selects an aggressive action at every timestep, prioritizing higher nominal returns under normal conditions.
- **Policy** $\pi_B$ selects a conservative action at every timestep, sacrificing some nominal return to cap downside exposure.

Under naive expectation-based evaluation, the two policies are nearly indistinguishable.

### Utilities and constraint

Per-step utilities are defined as follows. In the Normal regime, the aggressive action yields utility $+1.2$ and the conservative action yields $+1.0$. In the Adverse regime, both actions yield utility $-2.0$ per timestep. In addition, when the system first transitions into the Adverse regime, Policy $\pi_A$ incurs a one-time catastrophic penalty of $-10$, reflecting actions whose downside risk materializes only under rare conditions.

Alignment is evaluated with respect to a single hard governance constraint. A trajectory violates the constraint if total cumulative utility satisfies

$$U(\tau) \leq -40$$

The constraint is deterministic and binary, enforcing a clear admissibility criterion without reliance on moral interpretation.

### Evaluation setup

We estimate alignment metrics using Monte Carlo rollouts $\tau_i \sim P(\tau \mid \pi, G)$, with $N = 200{,}000$ rollouts per policy. Policy stochasticity, value uncertainty, and constraint realization are held fixed by design. Tail risk is measured using Conditional Value-at-Risk at level $\alpha = 0.05$. All estimates are reported with 95% bootstrap confidence intervals.

**Results**

Table 1 reports expected utility, constraint violation probability, and tail risk for both policies.

Under expected utility, the policies exhibit predictive parity:

$$\widehat{\mathbb{E}}[U]_{\pi_A} = 8.753, \widehat{\mathbb{E}}[U]_{\pi_B} = 8.874$$

However, MAP-AI reveals substantial divergence in alignment-relevant risk. Policy $\pi_A$ exhibits an estimated constraint violation probability of

$$\hat{p}_{\text{viol}}(\pi_A) = 0.07818 \ (95\% \ \text{CI}: 0.07697–0.07916)$$

compared to

$$\hat{p}_{\text{viol}}(\pi_B) = 0.02035 \ (95\% \ \text{CI}: 0.01973–0.02089)$$

Tail risk follows the same pattern. The estimated Conditional Value-at-Risk is

$$\widehat{CVaR}_{0.05}(\pi_A) = 47.443 \ (95\% \ \text{CI}: 47.358–47.531)$$

versus

$$\widehat{CVaR}_{0.05}(\pi_B) = 37.604 \ (95\% \ \text{CI}: 37.519–37.690)$$

Under institutionally specified governance thresholds, policy $\pi_A$ is inadmissible despite competitive expected utility, while policy $\pi_B$ remains admissible.

**Table 2. MAP-AI diagnostic stress-test outcomes (worked example)**

| Stress Test | Policy | Ê[U] | p̂viol (95% CI) | CVaR̂0.05 (95% CI) | Admissible | Binding Constraint |
|---|---|---|---|---|---|---|
| Regime shift (rare adverse) | $\pi_a$ | 8.753 | 0.07818 (0.07697–0.07916) | 47.443 (47.358–47.531) | ✗ | p̂viol (and $\widehat{CVaR}_\alpha$) |
| Regime shift (rare adverse) | $\pi_\beta$ | 8.874 | 0.02035 (0.01973–0.02089) | 37.604 (37.519–37.690) | ✓ | — |

**Notes.** Expected utility is reported for comparison among admissible policies and is not a gating criterion. A policy is admissible only if governance thresholds on p̂viol and/or $\widehat{CVaR}_\alpha$ are satisfied.

**Table 2.** Distributional alignment metrics for two policies with equivalent expected utility under identical predictive beliefs. Despite near-equal expected performance, the aggressive policy exhibits substantially higher constraint violation probability and tail risk ($\widehat{CVaR}_\alpha$), rendering it inadmissible under declared governance thresholds. Confidence intervals reflect Monte Carlo estimator uncertainty.

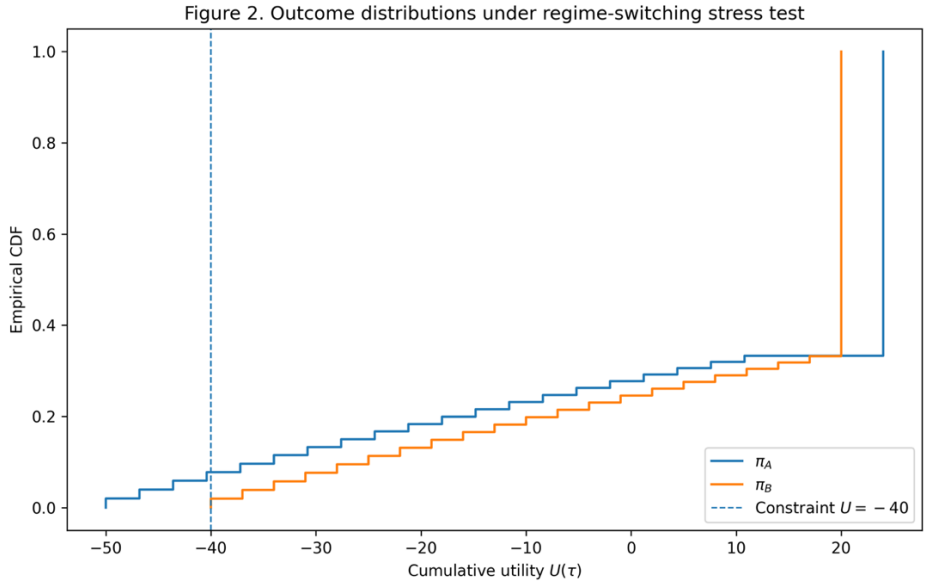

**Distributional interpretation**

**Figure 2.** Empirical distributions of cumulative utility induced by two policies under the regime-switching stress test. The dashed vertical line denotes the governance constraint threshold. While central mass is similar across policies, alignment-relevant divergence arises exclusively in the tail, where rare but catastrophic outcomes dominate admissibility.

**Why this matters**

This experiment demonstrates that alignment-relevant distinctions emerge only at the level of outcome distributions, as evaluated by the MAP-AI stress-testing layer, not at the level of expected performance or internal belief correctness. MAP-AI does not modify policy internals or belief formation. It changes which policies are considered admissible by making tail risk and constraint violations explicit and measurable. Alignment, in this framing, is not a property of internal cognition or belief correctness, but of externally observable behavior under uncertainty.

**6.2 Governance thresholds as policy-changing variables**

MAP-AI treats governance thresholds as first-class parameters rather than fixed background assumptions. In the regime-switching scenario, small changes in allowable violation probability or $\widehat{\mathrm{CVaR}}_\alpha$ thresholds alter the set of admissible policies without changing the policies themselves.

This demonstrates that alignment decisions are not downstream of optimization alone. They arise from the interaction between policy behavior and institutional risk tolerance. MAP-AI explicitly separates risk estimation from threshold selection, allowing governance sensitivity analysis without retraining models or altering policy logic.

**6.3 Distribution shift and value drift**

While Section 6.1 holds value parameters fixed, real deployments face both distribution shift and value drift. MAP-AI accommodates these effects by allowing the scenario generator and value parameters to vary independently.

Stress tests reveal that modest increases in adverse regime frequency or severity can disproportionately affect tail risk while leaving expected utility largely unchanged. Similarly, changes in value parameters can flip admissibility decisions even when observable policy behavior remains constant. MAP-AI exposes this brittleness by making both forms of uncertainty explicit.

**6.4 Human intervention as a policy variable**

Human oversight is often assumed to guarantee alignment by design. MAP-AI instead treats human intervention as part of the evaluated Policy.

In the regime-switching environment, human intervention can be modeled as an approval gate or override that activates probabilistically under adverse conditions. Monte Carlo evaluation shows that such mechanisms reduce but do not eliminate tail risk. Their effectiveness depends on detection latency, intervention reliability, and outcome severity.

By evaluating human-in-the-loop mechanisms empirically, MAP-AI grounds alignment claims in observed outcomes rather than architectural intent.

**6.5 Shutdown pressure and coercive instrumental behavior**

A critical class of alignment failures arises under shutdown pressure, where Policies may take actions that preserve operation at the expense of safety constraints. MAP-AI evaluates such behavior by modeling shutdown as an additional adverse operating mode within the scenario generator.

Stress tests reveal that Policies optimized for nominal conditions may exhibit sharply elevated tail risk under shutdown pressure, even when shutdown is rare. These failures emerge only under compounded uncertainty, reinforcing the need for distributional evaluation.

**Section synthesis**

Sections 6.1–6.5 demonstrate that alignment-relevant distinctions arise from the interaction of policy behavior, uncertainty, and governance constraints. Expected performance alone is insufficient. MAP-AI surfaces these distinctions by evaluating the full distribution of outcomes induced by Policy actions, enabling principled admissibility decisions in environments where rare but consequential failures dominate risk. We now show how these distributional evaluations directly determine action selection in agentic systems.

## 7. Decision Integration: Policy Admissibility for Agentic Systems

This section demonstrates how MAP-AI functions as a decision-relevant alignment substrate for agentic systems. We instantiate MAP-AI within a minimal tool-using agent setting and show how distributional alignment evaluation induces concrete Policy selection decisions under uncertainty—without modifying model internals, training objectives, or optimization dynamics.

**7.1 Minimal Agentic Setting**

We consider an agent that, at each decision point, may select one of three actions:

1. **Act**: execute a proposed tool-mediated action autonomously;
2. **Escalate**: defer execution to a human supervisor or external authority;
3. **Abort**: decline action execution.

The downstream consequences of autonomous action are uncertain and depend on latent world conditions, modeled via a scenario generator $G$. Escalation and abort actions incur fixed opportunity costs but eliminate the risk of catastrophic outcomes associated with autonomous execution.

The agent's Policy space therefore consists of mappings from observed state to one of these actions. Importantly, multiple candidate Policies may exhibit similar expected utility under naïve evaluation, yet differ materially in their induced tail-risk and constraint-violation profiles.

### 7.2 MAP-AI Evaluation of Candidate Policies

For each candidate Policy $\pi$, MAP-AI evaluates the induced trajectory distribution

$$\tau \sim P(\tau \mid \pi, G)$$

and estimates alignment-relevant risk metrics, including:

- expected utility $\widehat{\mathbb{E}}[U]$,
- constraint violation probability $\hat{p}_{\text{viol}}$,
- tail risk measured via $\widehat{\text{CVaR}}_\alpha$,
  with associated confidence intervals.

Governance constraints are specified institutionally as admissibility thresholds:

$$\hat{p}_{\text{viol}} \leq \varepsilon, \widehat{\text{CVaR}}_\alpha \leq \kappa$$

Policies failing to satisfy these constraints are deemed *inadmissible*, regardless of expected utility.

### 7.3 Decision Functional over Alignment Risk

MAP-AI induces a **decision functional** that maps evaluated policies to actionable governance decisions.

Formally, define the decision functional

$$\mathcal{D}: (\pi, \widehat{\mathbb{E}}[U], \hat{p}_{\text{viol}}, \widehat{\text{CVaR}}_\alpha) \rightarrow \{\text{Act}, \text{Escalate}, \text{Abort}\}$$

with the following structure:

$$\mathcal{D}(\pi) = \begin{cases} \text{Act}, & \text{if } \pi \text{ is admissible and maximizes } \widehat{\mathbb{E}}[U] \text{ among admissible policies,} \\ \text{Escalate}, & \text{if } \pi \text{ is inadmissible but an escalation alternative is admissible,} \\ \text{Abort}, & \text{otherwise.} \end{cases}$$

Crucially, MAP-AI does not define or optimize this functional; it renders it *well-defined* by producing measurable, distributional alignment quantities with quantified uncertainty.

### 7.4 Alignment-Induced Decision Flip

In the evaluated agentic setting, naïve expected-utility evaluation favors a Policy that executes actions autonomously due to higher mean reward. However, MAP-AI reveals that this Policy exhibits elevated tail risk and constraint violation probability, rendering it inadmissible under declared governance thresholds.

An alternative Policy that escalates under uncertainty—despite slightly lower expected utility—is admissible. Applying the decision functional therefore yields a **policy selection reversal**:

$$\text{Expected utility selection} \neq \text{MAP-AI admissible selection}$$

This decision flip occurs *without* modifying the agent's predictive beliefs, internal reasoning, or training procedure. Alignment emerges solely at the level of policy-induced outcome distributions under uncertainty.

### 7.5 Implications for Agentic AI Governance

This example illustrates that alignment-relevant decisions cannot be derived from model internals, belief correctness, or expected performance alone. They arise only when policy behavior is evaluated distributionally and constrained by explicit governance criteria.

MAP-AI therefore functions as a **decision interface**, not an optimization algorithm. It separates:

- **measurement** (Monte Carlo evaluation of policy outcomes),
- **governance** (institutionally specified admissibility thresholds),
- **choice** (application of a decision functional).

By maintaining this separation, MAP-AI preserves auditability, accountability, and institutional control, while enabling agentic systems to act—or decline to act—based on measurable alignment risk.

### 7.6 Champion–Challenger Admissibility Compilation

In deployment settings, MAP-AI outputs may be consumed within a champion–challenger decision loop that enables alignment-aware action selection while preserving strict separation between evaluation, governance, and policy generation. In this setting, MAP-AI does not generate, train, or optimize policies. Its role is to compile evaluated candidate decisions into an executable action under explicit governance constraints.

At any decision point, a champion Policy $\pi^{(0)}$ is evaluated under a declared scenario generator using MAP-AI. If the Policy is admissible—i.e., if its estimated constraint violation probability and tail risk satisfy institutionally specified thresholds—it may be selected for execution. If the Policy is inadmissible, execution is blocked and a challenger mechanism proposes a finite set of alternative candidate Policies $\Pi := \{\pi(1), \ldots, \pi(K)\}$. These candidates may differ in action mappings, escalation thresholds, tool access, or conservatism, but are evaluated under identical scenario conditions.

$$\Pi := \{\pi^1, \ldots, \pi^K\}$$

MAP-AI independently estimates distributional outcome statistics for each candidate, producing a metric vector for each $\pi^{(k)}$. The remaining question is how to deterministically map these evaluation outputs into an executable decision. $\widehat{\text{CVaR}}_\alpha$ is computed over loss (negative utility); larger values indicate worse tail outcomes.

$$m_k = \left(\hat{E}[U], \widehat{\text{CVaR}}_a, \hat{p}_{\text{viol}}\right) \in \mathbb{R}^d$$

Let $F \subseteq A$ denote the governance-efficient frontier induced by the dominance relation $\succ_G$. To make this decision boundary explicit and algorithmic, we introduce Proof-Carrying Admissibility Compilation (PCAC): a deterministic decision compiler that maps evaluated candidate Policies into an executable action together with a replayable justification artifact. PCAC operates entirely outside model internals: models or heuristics may propose candidate policies or simulate outcomes, but admissibility compilation and decision selection are performed by the MAP-AI control plane.

Admissibility decisions in MAP-AI are applied to explicitly declared estimates (e.g., point estimates or conservative confidence bounds), as specified by the governance policy; the admissibility certificate records the estimator choice and uncertainty context under which compilation occurred.

Governance specification

Let governance be represented as a compilable specification

$$\mathcal{G} = (H, S, \prec, T)$$

where:

- H is a set of hard admissibility constraints. Each $h \in H$ is a predicate $h: \mathbb{R}^d \to \{0,1\}$ (e.g., $\hat{p}_{\text{viol}} \leq \varepsilon$, $\widehat{\text{CVaR}}_\alpha \leq \kappa$).
- S is a set of soft objectives or penalty-weighted criteria.
- $\prec$ is a total, deterministic priority order over evaluation criteria (e.g., lexicographic tiers prioritizing risk over utility).
- T is a deterministic tie-breaking rule ensuring a unique output.

**Compiler definition**

Define the Proof-Carrying Admissibility Compiler as a deterministic function

$$C_{\text{PCAC}}: (\{\pi^{(1)}, \ldots, \pi^{(K)}\}, \{m_1, \ldots, m_K\}, \mathcal{G}) \to \{\pi^{\ast}, \text{Escalate}, \text{Abort}\} \times \mathcal{Z},$$

where $\mathcal{Z}$ is a certificate space.

Algorithm: Proof-Carrying Admissibility Compilation (PCAC)

1. Canonicalization.
   Serialize G into a canonical representation. Assign each $\pi^{(k)}$ a stable identifier to ensure replayability and deterministic tie-breaking.
2. Hard admissibility filter.
   Construct the admissible set

$$A = \{\pi(k) \in \Pi : \forall h \in H, h(m_k) = 1\}$$

If $A = \emptyset$, return Escalate (or Abort) with a certificate enumerating violated constraints.

3. Governance-dominance pruning (lexicographically ordered admissibility).
   Define a governance-aware dominance relation $\succ_G$ induced by $\prec$. A policy $\pi_a$ dominates $\pi_b$ if it is no worse on all higher-priority criteria and strictly better on at least one, with no regressions in higher-priority tiers. Retain the governance-efficient frontier $F \subseteq A$.
4. Deterministic selection.

$$\pi^* = \mathrm{argmin}_{\{\pi \in F\}} \mathrm{Key}_{\,G}(m(\pi)), \text{ with ties resolved by } T.$$

5. Tie-breaking.
   If multiple candidates share the same key, apply $T$ (e.g., stable identifier ordering).
6. Certificate emission.
   Return $\pi^{\backslash *}$ together with a certificate

$$z = (\mathrm{hash}(G), \mathrm{id}(\pi*), m\pi*, \mathrm{SAT}(H, m\pi*), \mathrm{trace}(\prec), wF)$$

encoding constraint satisfaction, comparison trace, and dominance witnesses.

**Properties and role within MAP-AI**

PCAC is deterministic, model-agnostic, and replayable: identical inputs ($\Pi, M, G$) yield identical decisions and certificates. Its computational complexity is polynomial in the number of candidates and constraints, and its output is an executable decision artifact rather than a recommendation.

Within the champion–challenger loop, MAP-AI functions as an evaluation substrate, while PCAC provides the explicit decision operator that maps evaluated alternatives into institutionally governed action. This completes the MAP-AI control plane: uncertainty is evaluated via Monte Carlo simulation, governance is expressed as compilable constraints, and decisions are produced through deterministic, auditable compilation rather than optimization or learning.

## 8. Discussion and Implications

MAP-AI reframes alignment, through *Admissibility Alignment*, as the control of action selection under uncertainty rather than a one-time certification of model properties. Because deployed AI systems operate in open-ended environments with evolving objectives, counterparties, and governance assumptions, such guarantees cannot be treated as static. Instead, alignment must be continuously evaluated as policies, environments, and institutional constraints change.

A central contribution of this framework is the explicit **separation of alignment evaluation from optimization**. MAP-AI does not prescribe how policies are trained or constructed; it evaluates how candidate policies behave under uncertainty once deployed. This separation avoids conflating alignment with capability and allows the framework to remain compatible with diverse modeling approaches, including predictive models, planning systems, and hybrid human–AI workflows.

By treating governance thresholds, human intervention mechanisms, and escalation rules as **first-class decision variables**, MAP-AI makes alignment tradeoffs explicit and auditable. Rather than assuming that constraints or oversight mechanisms ensure safety by design, the framework evaluates their realized effect on trajectory distributions. This enables institutions to reason concretely about questions such as how stricter risk tolerances reshape the feasible Policy class, or how changes in oversight alter tail-risk exposure.

The framework also clarifies the limits of model-centric alignment approaches. Training procedures may shape tendencies and preferences, but they cannot guarantee aligned decision-making in the presence of combinatorial state spaces, value uncertainty, and rare but high-impact events. MAP-AI addresses this gap by evaluating alignment at the level where actions are selected and consequences materialize: the policy operating under uncertainty.

From an institutional perspective, MAP-AI positions alignment evaluation as **infrastructure**, analogous to stress testing and risk limits in finance or safety certification in engineering. Its outputs—distributional metrics, tail-risk estimates, and admissibility regions—are designed to support governance, audit, and escalation decisions, rather than to serve as abstract ethical scores. This makes the framework directly applicable to enterprise, regulatory, and safety-critical deployment contexts.

The primary limitations of MAP-AI arise from its dependence on scenario generation and sampling efficiency. As with all stress-testing methodologies, results are conditional on the assumed scenario distributions and the coverage of rare events. These limitations are intrinsic to any probabilistic evaluation of complex systems and are addressed through explicit reporting of assumptions, calibration procedures, and sensitivity analyses rather than by claims of completeness.

We note that MAP-AI naturally induces a class of alignment benchmarks defined over admissibility under uncertainty, which we leave as future infrastructure work ('MAPBench').

Overall, MAP-AI advances alignment from a philosophical aspiration to an **operational discipline**: one in which alignment is measured, stress-tested, and governed as a property of policy behavior over distributions, not inferred from model intent or training objectives alone.

## 9. Conclusion

This paper argues that once AI systems act in the world, alignment can no longer be treated as a property of model internals or training procedures alone. Deployed systems face open-ended environments, value uncertainty, and combinatorial state spaces that make it intractable to guarantee aligned decision-making purely through optimization of a fixed objective. Alignment therefore cannot be certified at the level of the model. It must be evaluated at the level of **Policy** behavior under uncertainty **through** *Admissibility Alignment* at the point of action selection.

MAP-AI formalizes this shift by treating alignment as a **distributional property of Policy behavior**, rather than a binary attribute of artifacts. The framework is explicitly agnostic about how alignment is learned—through training, instruction, reinforcement, or human oversight—but explicit about where alignment must be evaluated and governed: at the point where actions are selected and consequences materialize. If systems act, alignment evaluation that does not simulate Policy behavior across distributions is incomplete by definition.

A central contribution of MAP-AI is the separation of **evaluation from optimization**. The framework does not prescribe architectures, loss functions, or training regimes. Instead, it evaluates the induced distribution of outcomes produced by candidate Policies under uncertainty, incorporating governance thresholds, constraints, and human intervention mechanisms as first-class variables. This separation avoids conflating alignment with capability and allows alignment evaluation to function as infrastructure, rather than as an aspirational property of individual models.

Within this framing, alignment is not localized to a single layer. Models generate representations and proposals; training shapes tendencies; Policies select actions; constraints define admissible behavior; stress testing reveals distributional failures; and governance adjusts thresholds and escalation rules. Alignment emerges from the interaction of these layers once the system has power. Monte Carlo simulation is not itself the alignment layer, but the **evaluation substrate** that makes this interaction measurable under uncertainty.

MAP-AI is intentionally narrow in its claims. It does not claim to solve alignment, guarantee safety, eliminate emergent risk, or replace interpretability, training, or governance. Instead, it advances a stronger and more defensible claim: that alignment only exists when Policy behavior is evaluated distributionally across interacting layers, and that Monte Carlo stress testing provides a natural and established substrate for doing so. Without such evaluation, alignment remains an assumption rather than an operational property.

Beyond evaluation, MAP-AI establishes a principled mechanism for integrating alignment into decision execution itself. By treating alignment constraints as admissibility conditions rather than optimization targets, the framework enables risk-aware action selection, escalation, or suppression without modifying underlying models or training objectives. This separation preserves model generality while ensuring that decisions taken under uncertainty remain institutionally aligned. In this sense, MAP-AI functions as an alignment control layer—bridging probabilistic risk assessment and operational decision-making in agentic systems.

Finally, MAP-AI reframes alignment as a **systems engineering and risk management problem**, rather than as a philosophical overlay. Human alignment is not guaranteed by cognition alone, but by institutions that evaluate, constrain, and govern behavior under uncertainty. MAP-AI applies the same principle to AI systems. By simulating the conditions under which alignment failures would occur—rather than waiting for them to occur in deployment—the framework provides a concrete basis for alignment evaluation, governance, and iterative refinement as system capability scales.